\documentclass[fleqn,10pt,twocolumn]{style-class}

\usepackage{xurl}
\usepackage[square,sort,comma,numbers]{natbib}

\title{Motion Illusions Generated Using Predictive Neural Networks Also Fool Humans}

\author{Lana Sinapayen${}^{1,2}$ and Eiji Watanabe${}^{2}$}
\speaker{Lana Sinapayen}

\affils{${}^{1}$Corresponding author, Sony Computer Science Laboratories, Kyoto Laboratory, Japan\\
(Tel: +81-75-456-1001; E-mail: lana.sinapayen@gmail.com)\\
${}^{2}$Laboratory of Neurophysiology, National Institute for Basic Biology, Okazaki, Japan\\
(Tel: +81-564-59-5595; E-mail: eiji@nibb.ac.jp )\\
}
\abstract{%
Why do we sometimes perceive static images as if they were moving? Visual motion illusions enjoy a sustained popularity, yet there is no definitive answer to the question of why they work. Here we present evidence in favor of the hypothesis that illusory motion is a side effect of the predictive abilities of the brain.  We present a generative model, the Evolutionary Illusion GENerator (EIGen), that creates new visual motion illusions based on a video predictive neural network. We confirm that the constructed illusions are effective on human participants through a psychometric survey.
Our results support the hypothesis that illusory motion might be the consequence of perceiving the brain's own predictions rather than perceiving raw visual input from the eyes. The philosophical motivation of this paper is to call attention to the untapped potential of 
``motivated failures", ways for artificial systems to fail as biological systems fail, as a worthy outlet for Artificial Intelligence and Artificial Life research.\\~\\
\textbf{Significance statement:} This manuscript presents evidence supporting the hypothesis that the perception of illusory motion in static images may be the result of the brain perceiving its own predicted visual input. The method proposed to support this hypothesis is the most innovative contribution of the manuscript, in its focus on validating an artificial model of perception by ensuring that the model's failures align with biological failures.\\~\\
\textbf{Classification:} Biological Sciences, Psychological and Cognitive Sciences}


\begin{document}

\maketitle


\section{Introduction}

\subsection{Artificial Perception}

The human perceptual system falls into the category of ``complex systems", where many parameters influence each other in ways that are difficult to disentangle. Complex systems can fail in ways that are exceptionally rich in information, and studying these failures teaches us more about those systems than we would learn by only observing successes~\cite{sinapayen2021perspective}. This is recognized in the Artificial Intelligence (AI) field the point that the discovery of ``adversarial" inputs (images, text prompts) are taken as evidence that most current AI models do not have enough in common with the human brain to be good models of perception~\citep{dujmovic2020adversarial, wichmann2023deep}. To help exploiting failure cases in the search to understand complex systems, we advocate for an approach that focuses on replicating biological failures in artificial systems. The approach follows these steps: 1. Document a failure (e.g. sudden loss of performance) in the biological system of interest; 2. Replicate the transition from failure to success in an artificial system; 3. Document novel (i.e. as yet not reported in the biological system) failures in the artificial system; 4. Verify that the novel artificial failures replicate in the biological system. An artificial system in which all steps are validated has an increased likelihood of sharing fundamental functions with the biological system, rather than displaying shallow resemblances. Applied to the field of perception studies, we call this approach ``Artificial Perception"~\cite{artPerc}. 

\begin{figure}[t]
    \centering
    \includegraphics[width=1.0\linewidth]{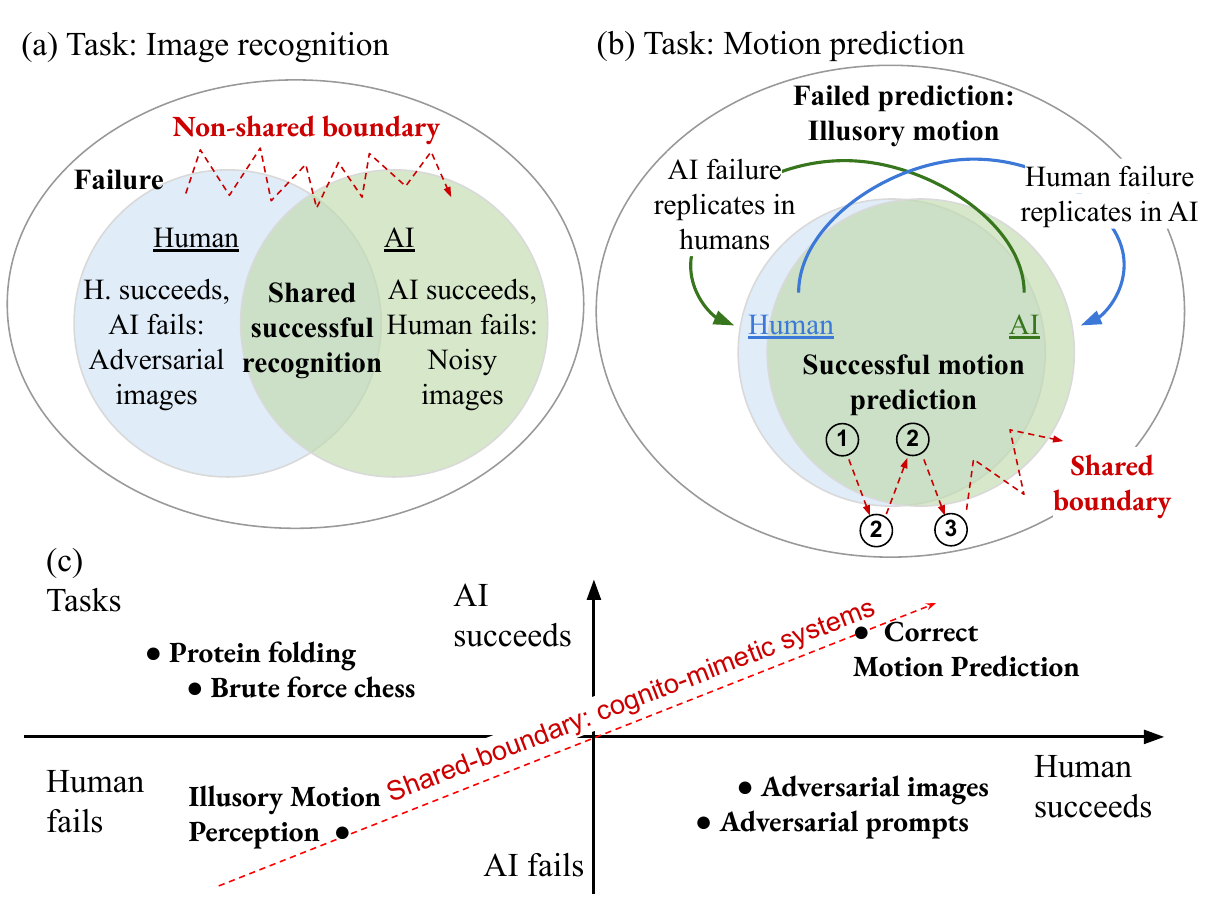}
    \caption{\textbf{The Artificial Perception Approach.} (a) Most AI models, for example here for image recognition, fail in ways that are utterly different from human failures: they do not work the same and therefore do not fail the same. (b) In this paper we use a 4-step method to show that a predictive AI model shares the same transition between failure and success as humans do, suggesting that the two systems also share working principles (in this case, predictive coding): they fail the same, and therefore may work the same. (c) The same approach, used for various tasks, would reveal other AI models that sit on the same failure boundary as humans, and therefore share working principles: cognito-mimetic systems.}
    \label{fig:c}
\end{figure}

\subsection{Illusions and Artificial Intelligence}

Input from our sensory organs can be heavily processed by sensory cells even before it reaches our brain, and various theories posit that most of this information is further processed by unconscious brain networks before reaching our awareness~\cite{velmans1991human,sklar2021non}.
For example, touch neurons in human skin have been found to compute the edge of objects before sending that information to the brain~\cite{pruszynski2014edge}; in the the visual system, the eye performs some complex computations of its own~\cite{deny2017multiplexed}, after which the brain itself performs high speed, unconscious computations before the information reaches our perception. This processed information, as well as our expectations, act as feedback that further modify incoming perceptual information~\cite{pylyshyn1999vision}. All of this intermediate data processing makes our perceptual system prone to errors of perception that are difficult to control or correct.

Illusions are some of the most spectacular, widespread, and enjoyable perceptual failures. Among the many existing definitions of ``illusion", this paper uses the following: an illusion is a perception that not only misrepresents reality, but can also be recognized by the observer through their other senses as a misrepresentation of reality. Furthermore, the knowledge that reality and perception do not match does not destroy the illusion~\cite{pylyshyn1999vision}. In summary, an illusion is a faulty perception that persists despite conscious knowledge and experience of the corresponding reality.
A number of Artificial Neural Network models have been found to respond to different types of visual illusions: color constancy illusions~\cite{gomez2019convolutional}, closure effects~\cite{kim2019neural}, the flash-lag effect~\cite{lotter2020neural}, the scintillating grid illusion~\cite{sun2021imagenet}, orientation illusions~\cite{benjamin2019shared}, length illusions~\cite{ward2019exploring}... 
A few papers tackle the task of synthesizing new illusions using AI: in~\cite{williams2016magic}, a version of Curry's paradox (``Missing square puzzle") is generated by an AI based on geometrical constraints and mathematical properties such as the number and size of jigsaw pieces. \cite{gomez2019synthesizing} use a more advanced neural network to generate brightness illusions for grey squares~\cite{gomez2019synthesizing} on generated backgrounds. Finally, in the last year, generative AI has and especially diffusion models have successfully been applied to the task of generating color illusions~\cite{gomez2025art} and bi- or multi-stable images~\cite{burgert2024diffusion,geng2024visual}: images that appear as one object but reveal a second object when the viewing conditions are changed (rotating, skewing, or shuffling the image).

\subsection{Motion Illusions}

This paper focuses on a type of visual illusion called ``illusory motion,"~\cite{fraser1979perception} where a single static image is perceived as if it were moving. In~\cite{fraser1979perception}, 75\% of participants reported perceiving illusory motion in the Fraser-Wilcox illusion, although the sensitivity to a given image seems to varies greatly. Fish, flies, monkeys, lions and cats have also been found to react to illusory motion~\cite{gori2014fish,tuthill2011neural,agrillo2015rhesus,regaiolli2019motion,baath2014cats}, although it is unclear whether they perceive it as a ``normal" motion or if they are aware that the motion is illusory.

Motion illusions, such as the famous Rotating Snakes Illusion~\cite{rotatingSnakes}, are typically created by trial and error. There is no consensus on why they work: is it because of eye movements~\cite{fermuller2010illusory}, asyncrhornous sensitivity to luminance~\cite{faubert1999peripheral}, or is it a static variation of the dynamic ``reverse phi" motion illusion~\cite{conway2005neural}? Although several theories have been advanced, only one offers a high level causal explanation: predictive coding. So far, we know that illusory motion seems to be influenced by eye saccades~\cite{faubert1999peripheral}, eccentricity and lighting~\cite{hisakata2008effects}, and often but not always require contrast changes. We also know that motion illusions perceived by humans also work on flies~\cite{agrochao2020mechanism}, macaques~\cite{conway2005neural}, domestic cats~\cite{baath2014cats}, lions~\cite{regaiolli2019motion} and fish~\cite{gori2014fish}. 
While the mechanisms could in theory be different in all these species, clearly there may also be some reason why animals all over the evolutionary tree are sensitive to the same perceptual trap.

\subsection{Predictive Coding}

The hypothesis of ``predictive coding" could provide much needed clues into the causal mechanisms of illusory motion. Predictive coding is the idea that a core function of brains is to predict their environment, and that learning mainly happens by minimizing errors between brain-generated predictions and reality~\cite{millidge2021predictive}. Some versions of the hypothesis postulate that the brain's most confident predictions suppress input from sensory organs, and experimental evidence supports that idea for the visual system~\cite{qin2023predictability}. Seeing motion illusions through the lens of predictive coding in the visual cortex~\cite{rao1999predictive}, it could be that the brain strongly predicts motion from patterns that ``usually are good predictors" of motion; this prediction suppresses sensory input, which would usually not be an issue if the prediction does not deviate much from reality. The illusion happens when a completely \textbf{static} pattern shares properties with these ``good predictors" of motion, causing the brain to predict and perceive motion over the veridical sensory input, leading to a gap with reality. A predisposition to predicting visual stimulus would be useful to various species, and could explain the prevalence of the illusion in the animal world.

Watanabe et al.~\cite{watanabe2018illusory} showed that a predictive deep neural network architecture called Prednet~\cite{lotter2016deep}, engineered based on predictive coding principles~\cite{rao1999predictive}, are tricked by motion illusions but not by modified patterns where the illusion was broken.
Prednet was trained using the First Person Interaction dataset (FPSI)~\cite{fpsi}, a first-person video of people going to an amusement park, filmed unrelated to any illusion research projects and a priori free of any purposefully introduced visual illusion. The task of the network was to predict future frames of the video.
The trained network predicts a rotating motion in the Rotating Snakes illusion, and no motion in a version of the illusion where the colors are swapped. Thus Step 1 and 2 of the Artificial Perception approach were cleared: 1. Documenting a failure (illusory perception of motion) in the biological system of interest, and 2. Replicating the transition from failure to success in an artificial system. Here we show that Step 3 (``Document novel failures in the artificial system") can also be cleared, and present evidence that Step 4 (``Verify that these failures replicate in the original biological system") is cleared as well, by collecting human data about the perception of our AI-generated illusions.
All 4 steps being cleared strongly suggest that Prednet is a closely cognito-mimetic system; and that, as the main common point between the biological and artificial system, predictive coding is a major cause of illusory motion perception.

In Section~\ref{methods} we describe the architecture of the illusion generator and its main parameters. In Section~\ref{results}, we present some of the generated illusions and their common points with existing motion illusions. We also present the results of data collection in humans. Finally in Section~\ref{discussion}, we explain some of the strengths and weaknesses of the presented results.

\section{Methods}
\label{methods}

\subsection{Survey Ethics}

All methods were carried out in accordance with relevant guidelines and regulations. The survey methods were approved by the ethics committee of the National Institutes of Natural Sciences, located in Japan. The permit number is EC01-078. Informed consent was obtained from all participants before starting the survey.

\subsection{Open Source Materials}

The architecture described here is named \textbf{E}volutionary \textbf{I}llusion \textbf{Gen}erator (EIGen), pronounced as in ``eigenvalue". It is available at
\url{https://github.com/LanaSina/evolutionary_illusion_generator}

The predictive models used as a plug-in evaluator are available at
\url{https://doi.org/10.6084/m9.figshare.13280120} (black and white) and
\url{https://figshare.com/articles/Sample_Weight_Model_Front_Psychol_15_March_2018_/11931222} (color).

The First Person Interaction Dataset dataset~\cite{fpsi}, now unavailable from the original source, is copied here: \url{https://figshare.com/articles/figure/FPSI_frames/7819574/1}) 

A collection of generated illusions is available at \url{https://figshare.com/articles/figure/EIGen_Visual_Illusions/16800013}

The participant questionnaire materials is published at \url{ https://app.gorilla.sc/openmaterials/1091991}.

The human perception data available at \url{https://doi.org/10.6084/m9.figshare.29977852}.

\begin{figure}[t]
    \centering
    \includegraphics[width=1.0\linewidth]{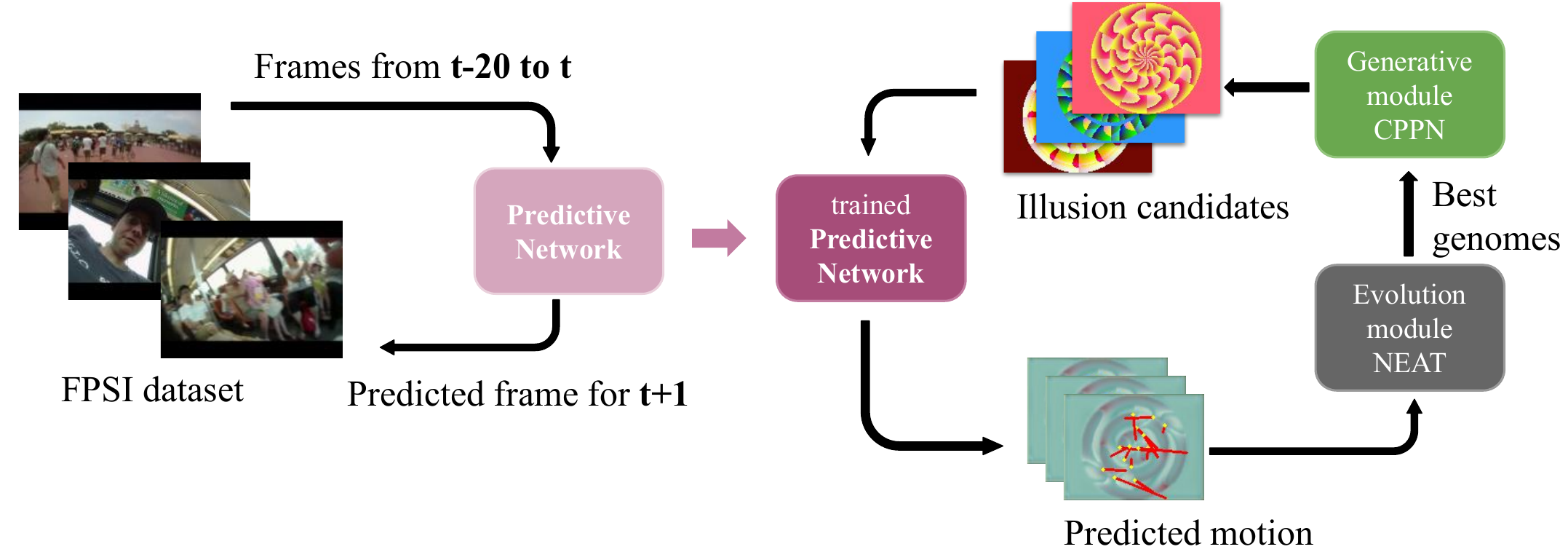}
    \caption{\textbf{The structure of EIGen.} A Predictive Neural Network (Prednet) is trained to predict video frames. It is then used in combination to optical flow calculation to rate the strength of illusory motion in images generated by Compositional Pattern-Producing networks (CPPN). The CPPNs are optimized by the NEAT evolutionary algorithm.}
    \label{fig:structure}
\end{figure}

\subsection{EIGen's Architecture}

EIGen is composed of two main modules (Fig.~\ref{fig:structure}): a generator and an evaluator. The evaluator is based on a fully pre-trained predictive neural network model and a visual flow calculator. The predictive model is (in advance) trained to predict video frames: taking a sequence of frames extracted from a video as input, the model has to output an image prediction as similar as possible as the next video frame~\cite{lotter2016deep}. We use two trained models: one trained on the original color video from the First Person Interaction Dataset dataset~\cite{fpsi} (now unavailable from the original source, but copied here \url{https://figshare.com/articles/figure/FPSI_frames/7819574/1}) and one trained on a black and white (greyscale) version of the dataset. Note that the results can be replicated with different datasets, as long as the model is shown to detect motion on well-known motion illusions.  In the evaluator module of EIGen, a single image is repeated 20 times to make a (static) input sequence. The output of the predictive model reveals the model's interpretation of whether the image is moving or not: for example, the output may be rotated compared to the original input, showing that the model interpreted the input as part of a rotation motion. As the input is just a sequence of one static image, the correct prediction would be a perfect copy of that image. Therefore, we take any predicted motion as ``illusory," analogous to the illusory motion that humans perceive in static images.
To estimate the direction and velocity of the predicted motion, we use the Lucas–Kanade method of calculating the optical flow between two images~\cite{lucas1981iterative}. This method compares the images and outputs the estimated origins and amplitudes of motion vectors originating from the first image and leading to the second image (see Fig.~\label{fig:illusions_bw} for example).

The generator module of EIGen uses Compositional Pattern-Producing Networks (CPPNs~\cite{stanley2007CPPN}) to generate images. As their name indicate, CPPNs generate patterns; these can be interpreted as images. The structure of CPPNs makes them well suited to artificial evolution. Here we use the NeuroEvolution of Augmenting Topologies (NEAT~\cite{stanley:ec02}, specifically its python implementation~\cite{neat-python}) evolutionary algorithm to optimize the CPPNs' parameters. 
The working principle of the NEAT algorithm is as follows: generate several ``genomes" (parameter sets) grouped into ``families", keeping the families as diverse as possible; iteratively mutate the genomes and keep the best genome in each family (according to a user-defined fitness function) for the next round of evolution. In other words, the competition is only between genomes in the same family, and each family can explore a different part of the parameter space. In EIGen, the parameters of the CPPNs are treated as genomes. The genomes are evaluated as follows: first, the CPNN genome produces an image. The evaluator module returns the group of motion vectors associated to this image. The group of vectors is then scored according to our fitness function. This score is used to rank the genomes against each other.

\subsection{Fitness function}

At first glance, it may seem like the best illusions would simply be the images with the biggest motion vectors: a big prediction error = a big illusion. In practice, we found that there are other conditions common to motion illusions (including the vast majority of human-created ones) that have a strong effect on observers. The best illusions combine these conditions: (a) numerous motion vectors, indicating many sources of illusory motion; (b) big motion vectors, indicating a strong illusory motion (technical caveat: huge motion vectors tend to be due to instabilities in the optical flow module); (c) vector orientation and sense closely aligned to the some of the neighboring vectors, indicating agreement in the predicted motion; (d) vectors with same direction but opposite sense to the some of the neighboring vectors, indicating contrast in the predicted motion (this condition is fullfilled by our choice of structure for the image: concentric circles with patterns alternating clockwise and counterclockwise directions). 

While some of these conditions are acknowledged in the literature, the reasons behind their importance has not been elucidated.

Through trial and error, we converge towards the following calculation to get a fitness score $fitness$ from a list of motion vectors \emph{list\_v}:

\begin{verbatim}

# Remove implausibly big vectors 
# from list_v
max_plausible_norm = 0.3
for v in list_v:
    if norm(v) > max_plausible_norm:
        list_v.remove(v)

# condition (a):
# there should be many vectors
if length(list_v) < 24
    fitness = 0
else
    # condition (c): vectors are aligned
    # if their variances x and y are low
    list_norm_v = normalize(list_v)
    s =  (1 - pow(var(list_norm_v[:,0]),2) 
    + (1 - pow(var(list_norm_v[:,1]),2)
    score_direction = s / 2
    
    # condition (b): ranks high if vectors
    # are big and aligned.
    mean_x = mean(abs(list_v[:,0])
    s = mean_x / max_plausible_norm
    variance = variance(norms(list_v))
    score_strength = 
        s * (1 - min(variance, 1))
    
    # higher weight on alignment
    # gives better results
    fitness = 0.7 * score_direction 
        + 0.3 * score_strength

\end{verbatim}

This ad hoc nature of the fitness function is discussed later in section~\ref{discussion}.

\subsection{Generating illusions}

We constrain the structure of the output images to circles with concentric bands of repeating patterns. The patterns of two neighboring bands are inverted to be alternatively oriented clockwise and counterclockwise (Fig.~\ref{fig:illusions_bw}). In itself (for example when filled with random colors) this structure is not sufficient to induce illusory motion, but over the course of evolution it makes it easier for the generator to fulfill conditions (c) and (d). It is similar to the structure often used in the Rotating Snakes illusion.
One generation of the genetic algorithm corresponds to one pass through each module: mutating genomes, and scoring the corresponding images. EIGen can run indefinitely; in this paper we stop it when the fitness reaches 0.8 out of 1.

\subsection{Parameters}
PredNet parameters: image width = 160, image height = 120; input length = 20 images; extension duration = 2 images. NEAT parameters: species size = 10 individuals, population = 5 species.

\subsection{Evaluation by human participants}

Ethics approval number EC01-078 by the National Institute of Natural Sciences, Tokyo, Japan. 

The dataset of illusions for human evaluation comprises human- and EIGen-created illusions.
For the human illusions, we chose the Rotating Snakes Illusion (as the strongest known and most studied illusion of this type), and the Fraser-Wilcox illusion~\cite{fraser1979perception} (as the first discovered illusion of this type, and as an illusion several times ``rediscovered" by EIGen).
For the EIGen-generated illusions, we choose two black and white images with rotating (tangential) vectors, two with with expanding (normal) vectors, two color images with expanding (normal) vectors, one control image (no motion vectors) and one image that happened to replicate a previously human-created illusion, the medaka illusion~\cite{medaka}. See Section[] for a discussion of existing illusions ``rediscovered" by EIGen.

Except for the control and medaka images, we take the first images where the fitness is above 0.8 (i.e. we do not select what we subjectively perceive as ``good" images).

For this total of 10 images, we gathered the responses of $N=293$ participants. We collect the data using the Gorilla Experiment Builder platform \url{www.gorilla.sc}~\cite{anwyl2020gorilla}. The full materials of the questionnaire are published at \url{ https://app.gorilla.sc/openmaterials/1091991}. Appendix X shows an example of evaluation questionnaire: participants are shown the image and asked whether they perceive any kind of motion; if yes, they are asked to rate the motion qualitatively (is the perceived motion rotational or radial) and quantitatively (how strong is the perceived motion). We also collect free comments. Data was collected on the 28th March 2025. 
Participants were recruited using the Prolific platform (\url{www.prolific.com)}) under the following conditions: must be fluent in english, use a computer (not a smartphone), corrected vision. Participants demographics in appendix.

\begin{figure}[t]
    \centering
    \includegraphics[width=1.0\linewidth]{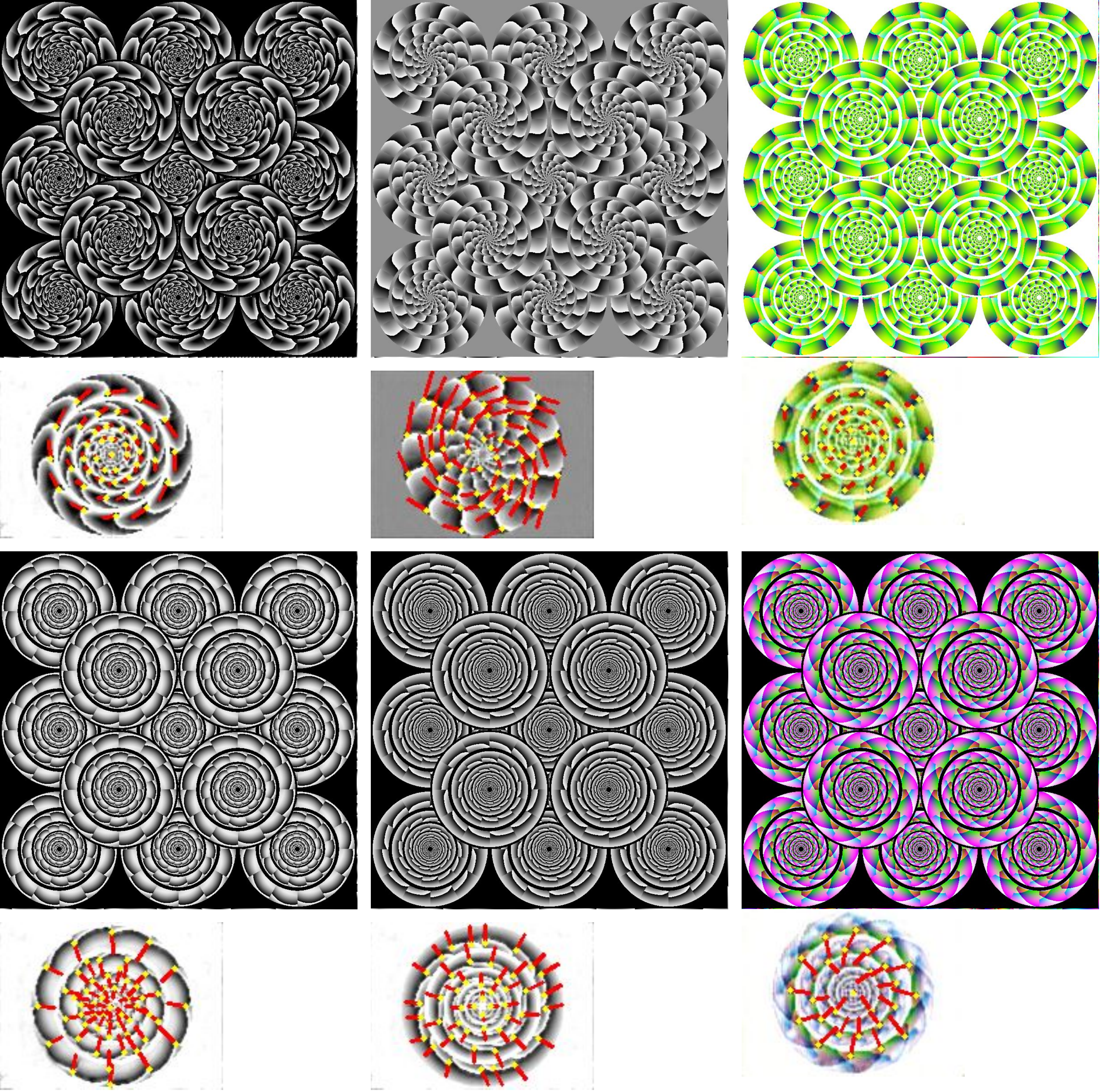}
    \caption{\textbf{Examples of greyscale and color illusions generated by EIGen.} The smaller image at the bottom of each illusion represents the motion predicted by the network. The vectors origins are marked as yellow dots; the amplitude is multiplied by 60 for easy visualization. These images are handpicked for illustration purposes; the images that were used for the human perception experiment are in the Appendix.}
    \label{fig:illusions_all}
\end{figure}

\begin{figure}[t]
    \centering
    \includegraphics[width=1.0\linewidth]{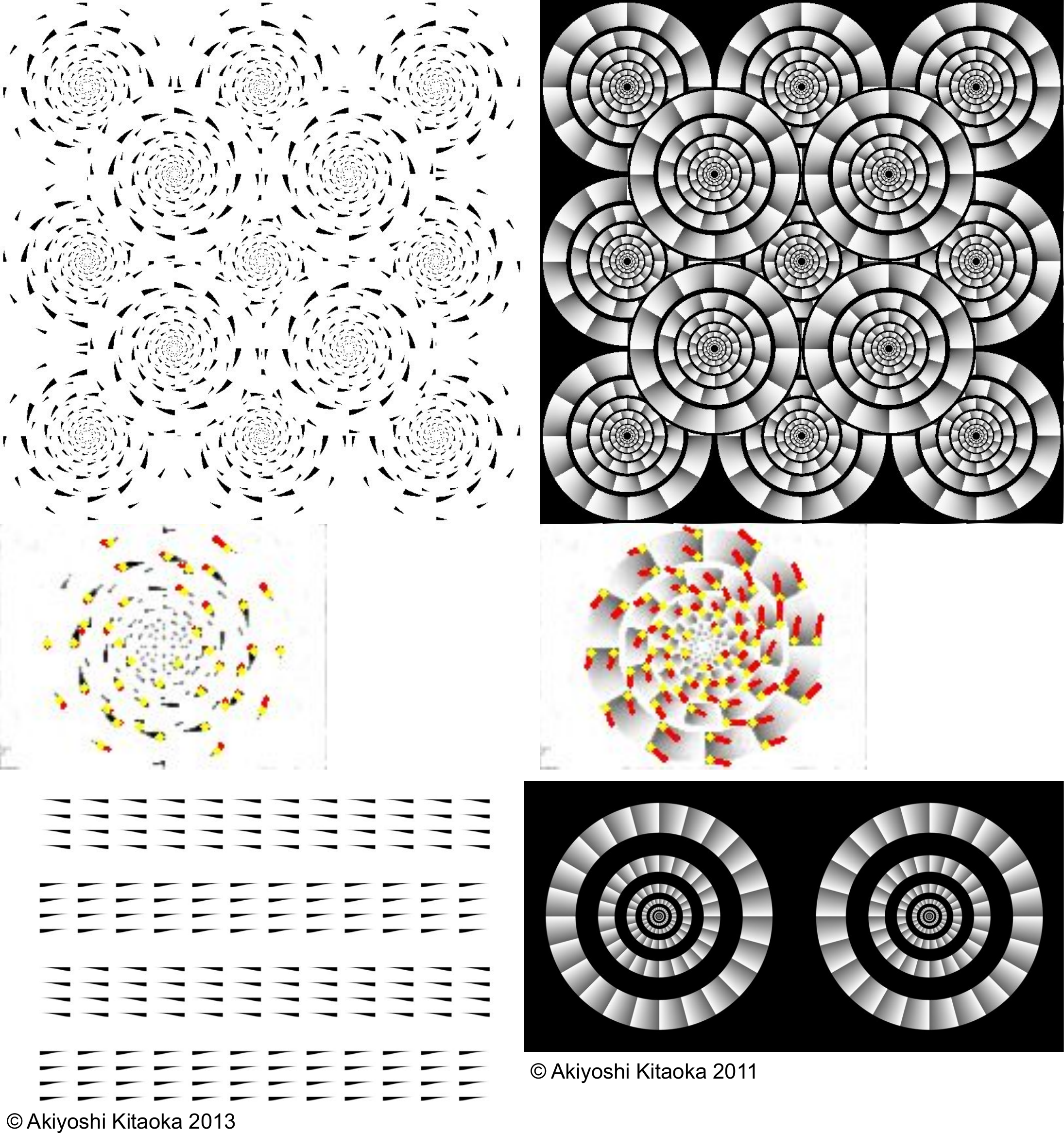}
    \caption{\textbf{Generated illusions replicating existing human-designed illusions.} The top left image is similar to A. Kitaoka's `medaka school` illusion (bottom left)~\cite{medaka}, except that Kitaoka's illusion is linear. EIGen invariably converges to this solution when forced to use a binary output (pixels fully black or fully white). The motion predicted by the network is extremely small, and accordingly one of the author does not perceive motion on either the EIGen-generated illusion nor Kitaoka's illusion. The top right image is a close replication of Kitaoka's 2011 iteration (bottom right,~\cite{kitaokaFraser}) on the Fraser-Wilcox illusion~\cite{fraser1979perception}. This output is also frequent. Images reproduced with permission.}
    \label{fig:replications_bw}
\end{figure}

\begin{figure}[t]
    \centering
    \includegraphics[width=1.0\linewidth]{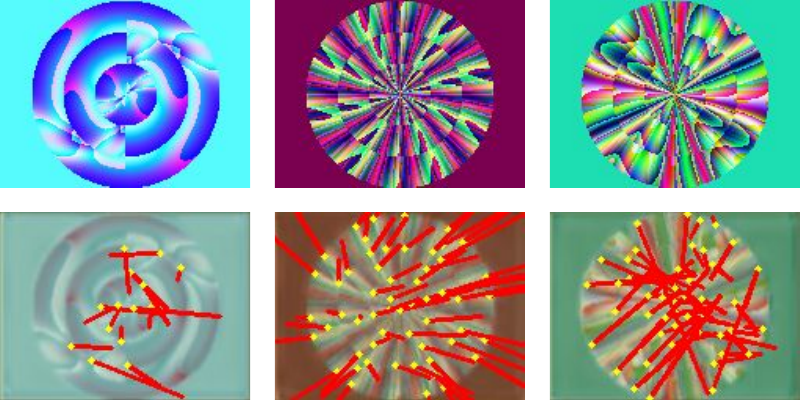}
    \caption{\textbf{Failed illusions.} These images did not produce illusory motion, even when combined through duplicated and mirroring structures as in other figures. The first image has few vectors. The second image has many vectors but their direction and position is unclear. Same for the last image.}
    \label{fig:failures}
\end{figure}

\section{Results}
\label{results}

\subsection{Generated images}

Common tips to best experience the illusory motion include using good lighting and slowly moving one's gaze around the image rather than focusing on one point.
Fig.~\ref{fig:illusions_bw} shows novel illusions obtained with the black and white and the color model (note that the actual images used for the human participant survey are in the Appendix), while Fig.~\ref{fig:replications_bw} presents illusions that happen to reproduce known illusions: the Fraser-Wilcox illusion, and the Medaka illusion. The fact that EIGen could ``rediscover" existing illusions, and predict motion in the expected orientation, is a hopeful sign of the agreement between the artificial and the biological data. 
We also present examples of failed illusions~(Fig.\ref{fig:failures}, where the model predicts a motion but no motion is perceived by the authors or other humans in an informal poll. These failed illusions illusions were all obtained with fitness functions that did not follow the 4 conditions outlined in the Methods section. This justifies using a fitness score that is not simply equal to the size of the visual flow vectors, but a more nuanced heuristic as described earlier.

\begin{figure}[t]
    \centering
    \includegraphics[width=1.0\linewidth]{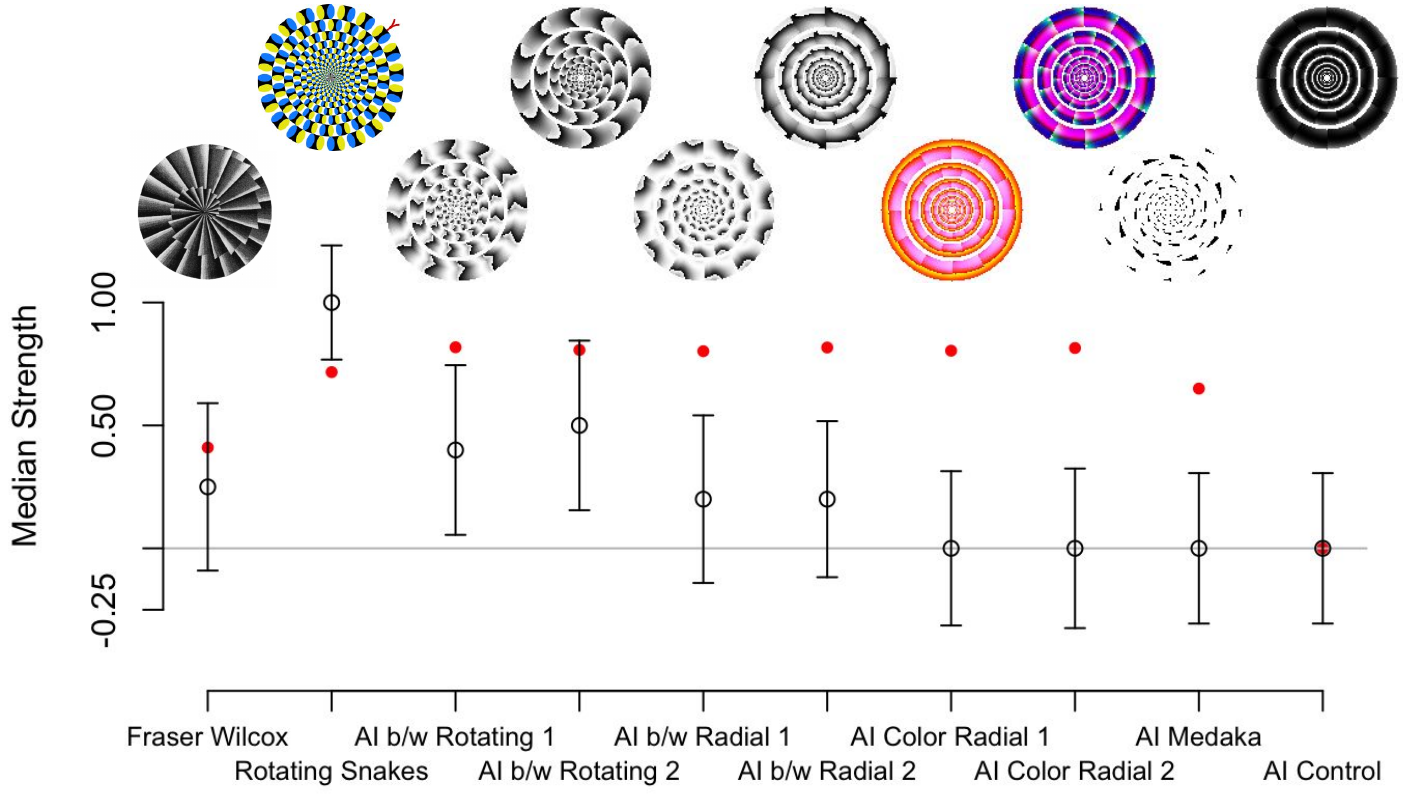}
    \caption{\textbf{Median human perception (black) and AI perception (red).} The insets represent part of each image for reference only; the complete images presented to the participants can be found in the Appendix. The AI-generated control, medaka, and color images all rank at 0, no motion perceived by most participants (detailed analysis of Fig~\ref{fig:histogram} reveals a more nuanced picture. The AI-generated black and white illusions fare better, especially for rotating illusory motion. The Rotating Snakes illusion remains undefeated.}
    \label{fig:medians}
\end{figure}

\begin{figure}[t]
    \centering
    \includegraphics[width=1.0\linewidth]{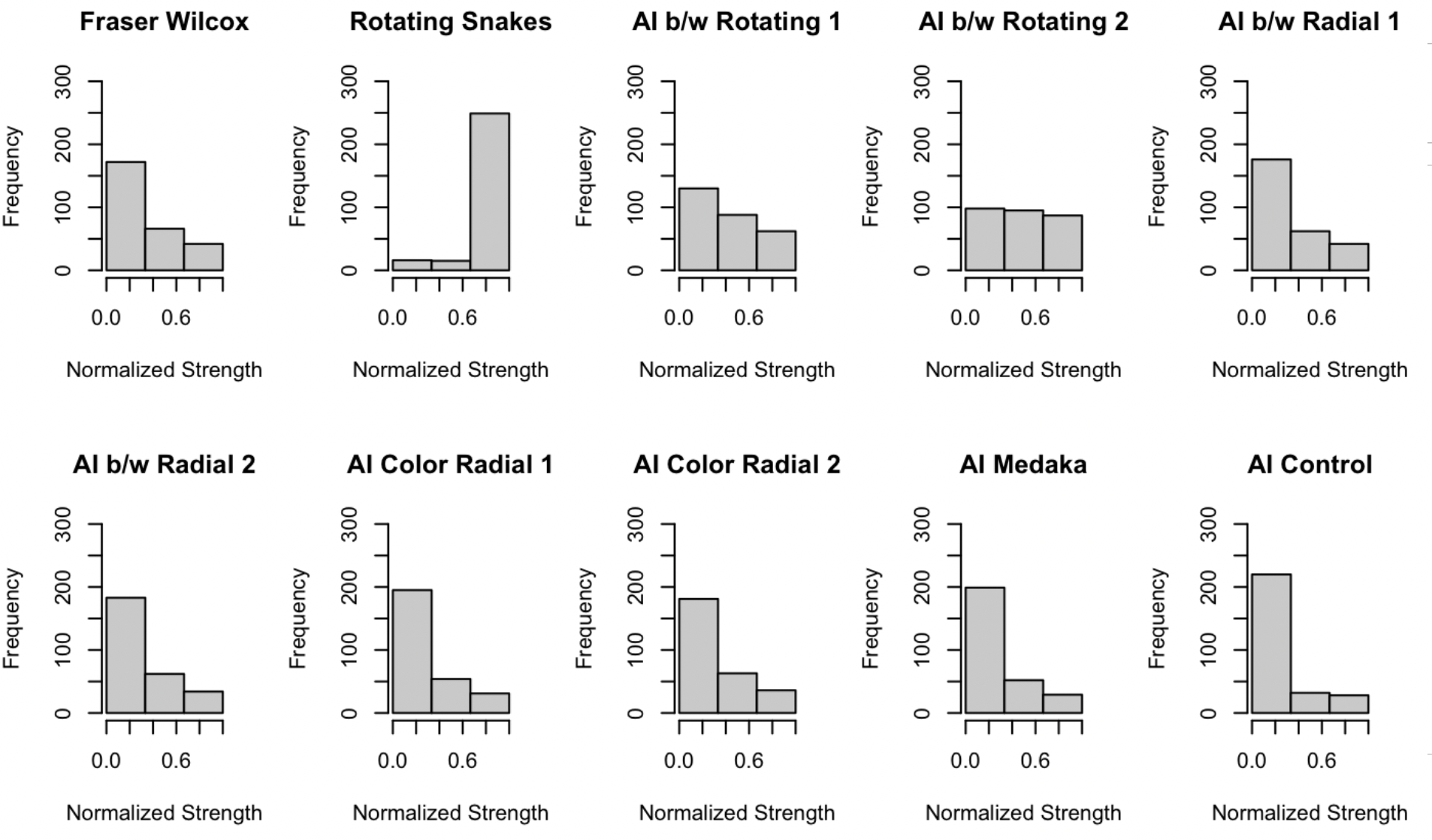}
    \caption{\textbf{Histogram of participants perception for each image.} The Rotating Snake and the control image are outliers: all answers 0 (control) or 3 (Rotating Snakes). Other images show the variability of perceptions between participants: for black and white rotating AI illusions, most people perceive some motion. Except for the control and the medaka, about 1/3 of participants perceive some degree of illusory motion in all other AI generated images.}
    \label{fig:histogram}
\end{figure}

\subsection{Quantitative Evaluation}

Participants rated the illusions from 0 (no illusion) to 5 (strong illusion). To minimize individual variation in the tendency to perceive or not perceive illusory motion, we normalize the results by dividing all ratings for a given participant by the strongest rating from this participant. This gives a score between 0 and 1 that we can compare with the fitness score of EIGen. 

Fig.~\ref{fig:medians} shows the median of human ratings (black with error bars) in contrast to the EIGen fitness rating of the same illusions. We use the median rather than the mean because the distribution of scores (Fig.~\ref{fig:histogram}) is not Gaussian. Within the human created illusions, the Rotating Snake illusion got the maximum score of 1, the strongest motion far above any other image. This is in agreement with the literature. The Fraser-Wilcox illusion was rated rather low with a strength 0.25.

Within the AI-created illusions, in general the median scores are not significantly different from each other, as evidenced by the error bars. Nevertheless, generally speaking the black and white gradient images with rotating motion were rated best (weaker than the Rotating Snakes but above the Fraser Wilcox illusion), followed by the the black and white gradient images with expanding/shrinking motion, and finally the color illusions, the rediscovered medaka, and the control were rated not moving with a score of 0.

The takeaway from these results is that they echo anecdotal evidence collected beforehand: the black and white model produces better illusions than the color model on par with the human-created black and white illusions; the Rotating Snakes trumps all other illusions. The difference between the EIGen fitness score and the human ratings show that our ad hoc fitness function does not capture the full nuance of human perception, although it is sufficient for producing black and white illusions. While out of scope for this paper, it would be interesting to search for the true fitness parameters matching the shape of the human perception distribution from these results, either from theory or from regression.

We also analyze the distribution of responses by coarse graining the human scores (bins of equal width) into 3 categories: 0 (no motion), 1 (some motion), 2 (strong motion). 
Fig.~\ref{fig:histogram} shows the histogram of responses for all illusions. The Rotating Snake and the control image are outliers, with dirac distributions: all answers 0 (control) or 3 (Rotating Snakes). Other images show more variance in the responses, highlighting the variability of perceptions between participants: for all of these images, some people do perceive illusory motion. ?At least half as many people perceive some type of motion as the number of people perceiving no motion.

\subsection{Qualitative Evaluation}

\begin{figure}[t]
    \centering
    \includegraphics[width=1.0\linewidth]{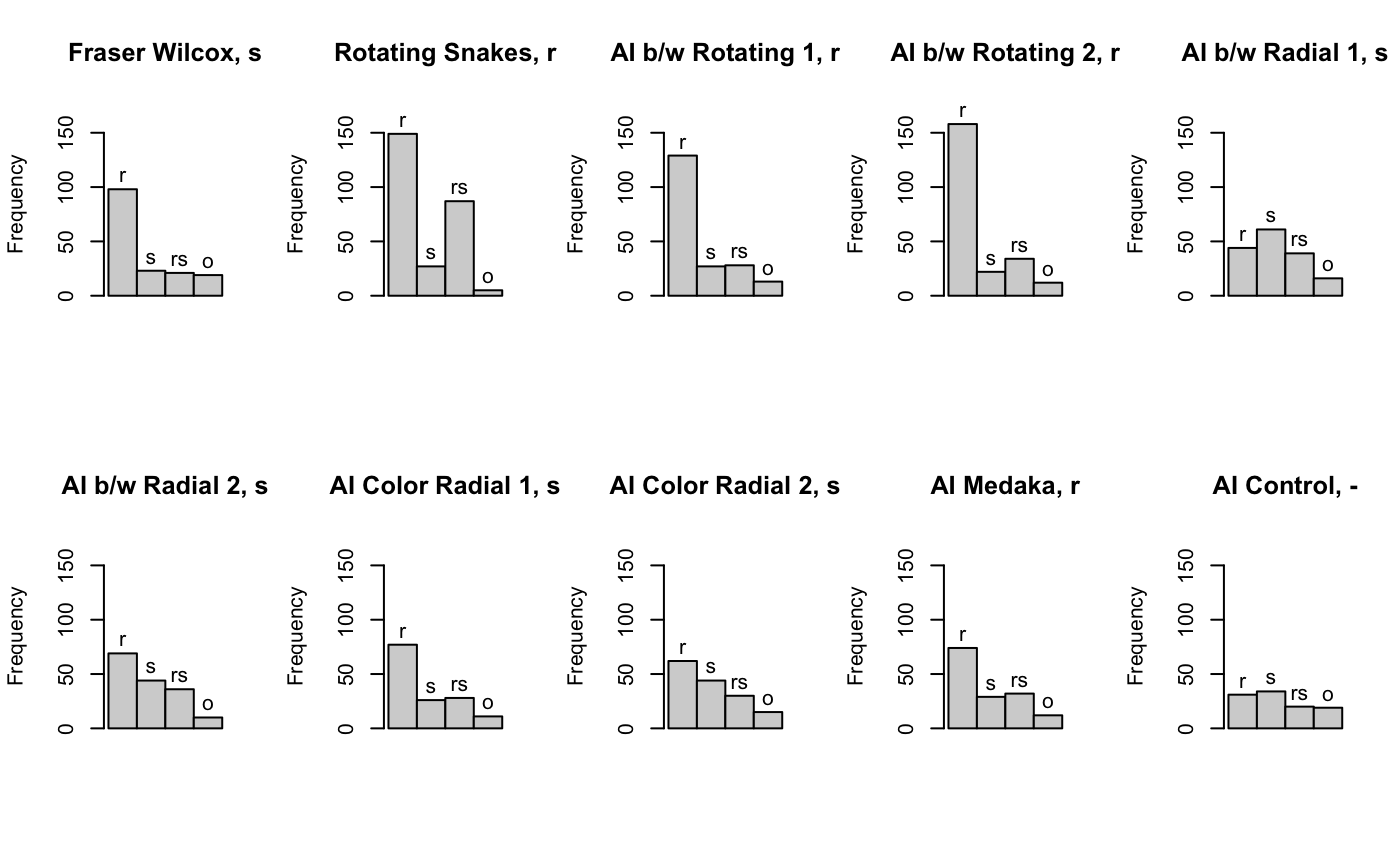}
    \caption{\textbf{What type of motion do participants perceive?} For each image, participants who reported perceiving illusory motion indicated the type of motion: r (rotating), s (shrinking/expanding), rs (both), o (other). The AI prediction is indicated on top of each histogram. Human participants to perceive mainly illusory rotations on all images (except the control), even when the AI model predicts shrinking/expanding motion. Only one image elicits a majority of shrinking/expanding illusory motion (this time in accordance with the AI prediction).}
    \label{fig:qual}
\end{figure}

Participants who indicated perceiving motion in an image were invited to classify the motion as either radial (expending/shrinking) or tangential (rotating). Fig.~\ref{fig:qual} shows that most of the perceived motion is ``rotating". There are notable exceptions: for the two EIGen generated image that were predicted by the network to be shrinking/expanding, human participants show less bias toward perceiving a rotating motion; one image is even primarily perceived to be shrinking/expanding as predicted.
The othe exception is the rotating snakes, which is perceived by most human participants to be rotating, but also by a significant number of participants to be both rotating and shrinking/expanding. Could this be part of the secret of the overwhelming strength of the Rotating Snakes?

Participants also had the possibility to leave free comments on each illusion. The comments revealed that many images were perceived as inducing several different types of illusions:
``"In addition to rotating, the circles also appear as a ``ghost image" in the white part of the webpage. The ghost images also rotate."; ``Vibrating/shaking"; ``Flickering"; ``A bit of a alternating darkening and lightening of the center of the circles."; ``it doesnt exavtly rotate, i can see the circles move as a whole but not spin".

As a whole, these results validate some observations made in our~\cite{sinapayen2021evolutionary} preprint, before collecting data: 
\begin{enumerate}
    \item EIGen can generate images that induce perceived motion in human observer
    \item The black and white model generate illusions that are stronger than the color model (the color illusions have barely any motion)
    \item The best of the NN illusions are far weaker than the best known human generated illusions
\end{enumerate}

\section{Discussion}
\label{discussion}

This paper presents a model, EIGen, that generates new visual illusions by coupling artificial neural networks and genetic algorithms. EIGen represents Step 3 and 4 of the Artificial Perception approach for illusory motion:  we document novel failures, in this case illusory motion in various images, in an artificial system; and we verify that the artificial failures replicate in the original biological system. We show that EIGen rediscovers existing illusions: this suggests that at least a subset of the generated illusions shares all characteristics of illusions that work on humans.
Weaknesses remain: first, we do not obtain illusions with the overwhelming effect of the Rotating Snakes. Second, we use an ad hoc fitness function, which may not correspond to the human scoring function, and makes it difficult to know if ``bad" illusions are bad because of inaccuracies in the prediction model or because of inaccuracies in the fitness function. Finally, we have no justification for 3 of the 4 conditions that we implemented in the fitness function. It makes sense that bigger motion vectors makes for stronger illusions, but  why is it better to have more vectors? Why do the vectors need both a level of agreement and a level of disagreement for the illusion to be perceived by humans, while the model predicts motion either way? Clearly, the model does not entirely overlap with human perception, as it produces motion vectors when these conditions are not satisfied. Nevertheless, it seems fair to say that this paper strengthens the idea that predictive coding is involved in the perception of illusory motion in humans, and that predictive neural networks share functional rather than superficial characteristics with human perception.

\section{Funding}
No funding besides the authors' affiliated institutions.

\newcommand{\newblock}{\ }
\bibliographystyle{IEEEtran}
\bibliography{bibliography} 

\clearpage

\section{APPENDIX}

\begin{figure}[ht!]
    \centering
    \includegraphics[width=1.0\linewidth]{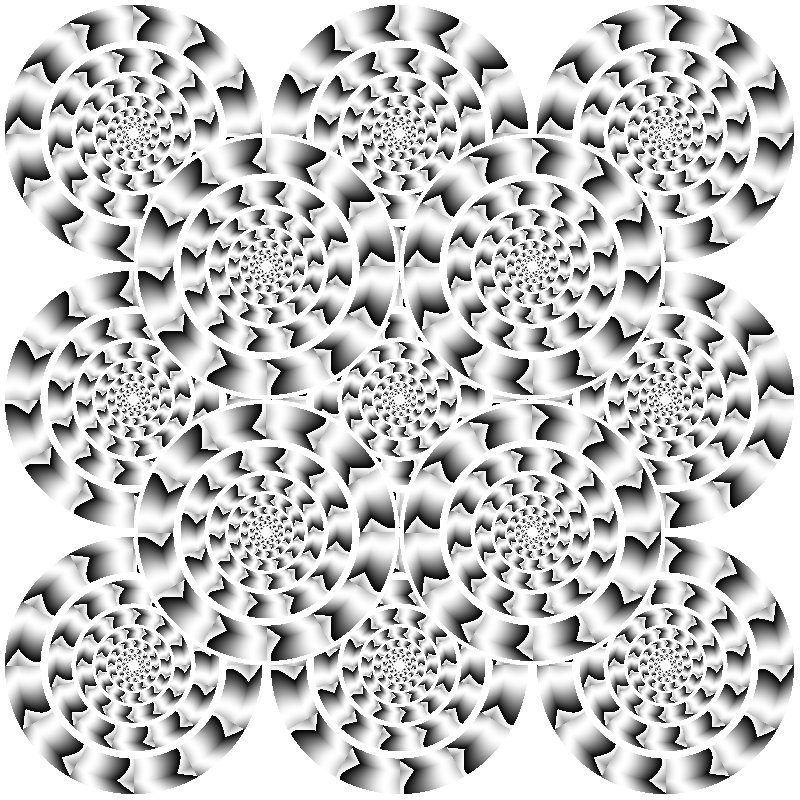}
    \includegraphics{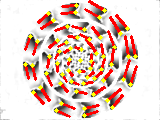}
    \caption{\textbf{EIGen-generated image} referred to as ``AI b/w Rotating 1" in the perceptual experiment, and its calculated visual flow.}
    \label{fig:app:1}
\end{figure}

\begin{figure}
    \centering
    \includegraphics[width=1.0\linewidth]{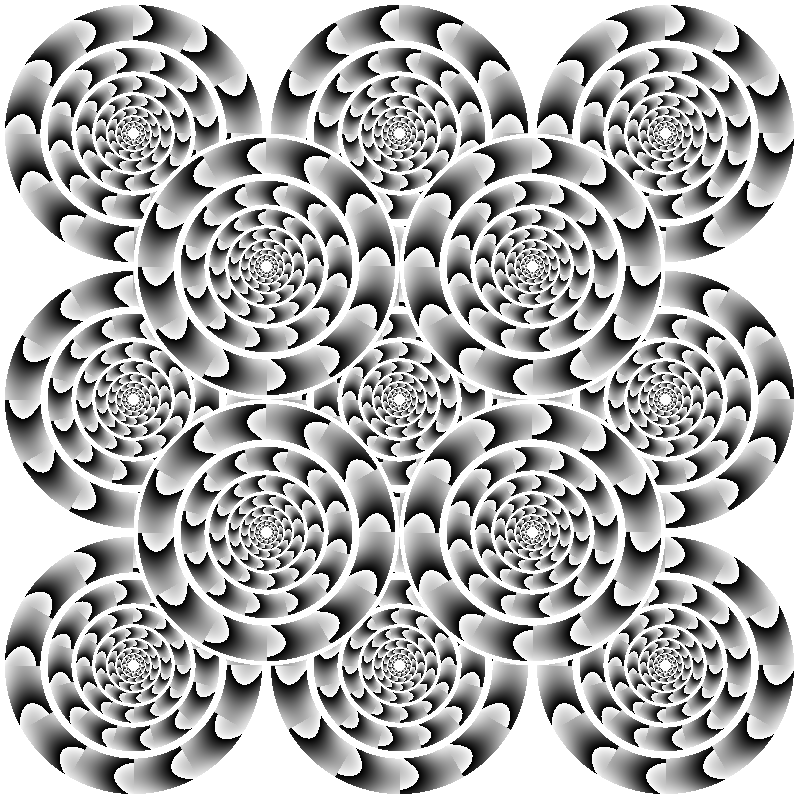}
    \includegraphics{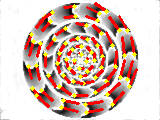}
     \caption{\textbf{EIGen-generated image} referred to as ``AI b/w Rotating 2" in the perceptual experiment, and its calculated visual flow.}
    \label{fig:app:2}
\end{figure}

\begin{figure}
    \centering
    \includegraphics[width=1.0\linewidth]{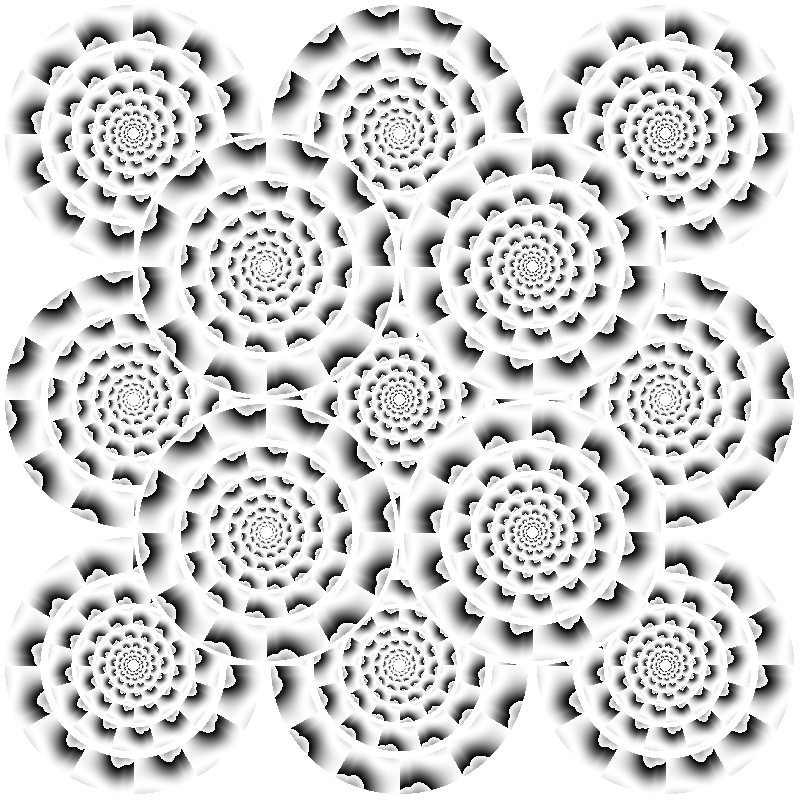}
    \includegraphics{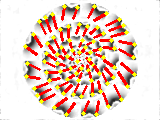}
     \caption{\textbf{EIGen-generated image} referred to as ``AI b/w Radial 1" in the perceptual experiment, and its calculated visual flow.}
    \label{fig:app:3}
\end{figure}

\begin{figure}
    \centering
    \includegraphics[width=0.9\linewidth]{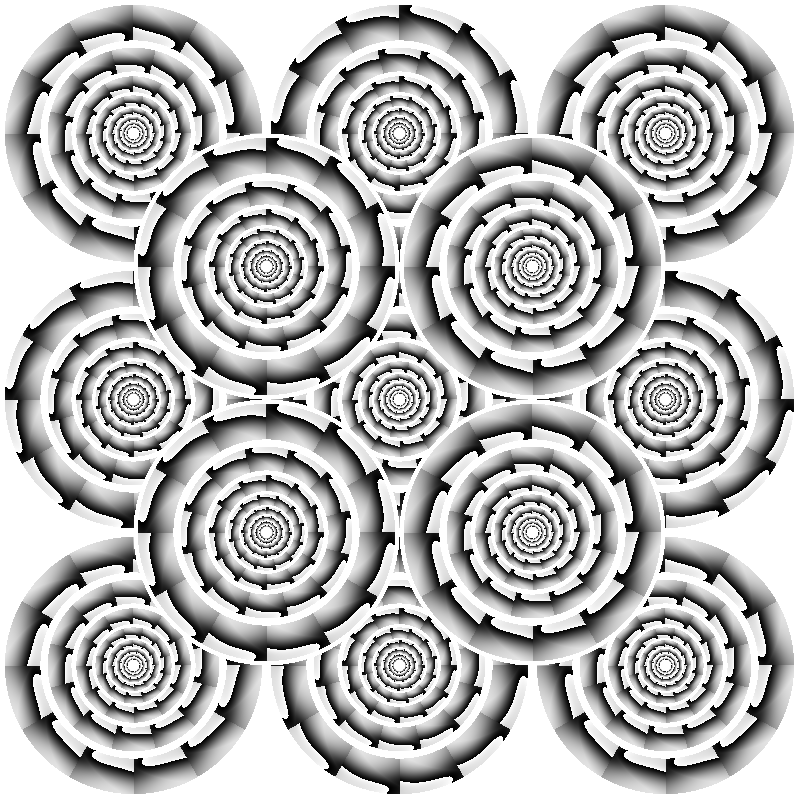}
    \includegraphics{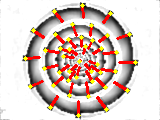}
     \caption{\textbf{EIGen-generated image} referred to as ``AI b/w Radial 2" in the perceptual experiment, and its calculated visual flow.}
    \label{fig:app:4}
\end{figure}

\begin{figure}
    \centering
    \includegraphics[width=1.0\linewidth]{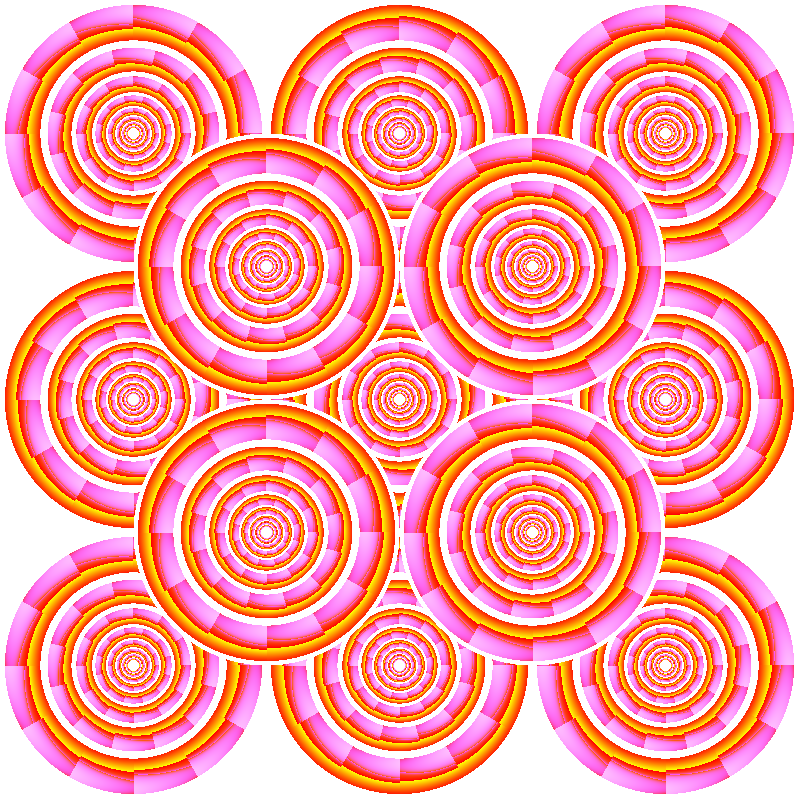}
    \includegraphics{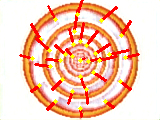}
     \caption{\textbf{EIGen-generated image} referred to as ``AI color Radial 1" in the perceptual experiment, and its calculated visual flow.}
    \label{fig:app:5}
\end{figure}

\begin{figure}
    \centering
    \includegraphics[width=1.0\linewidth]{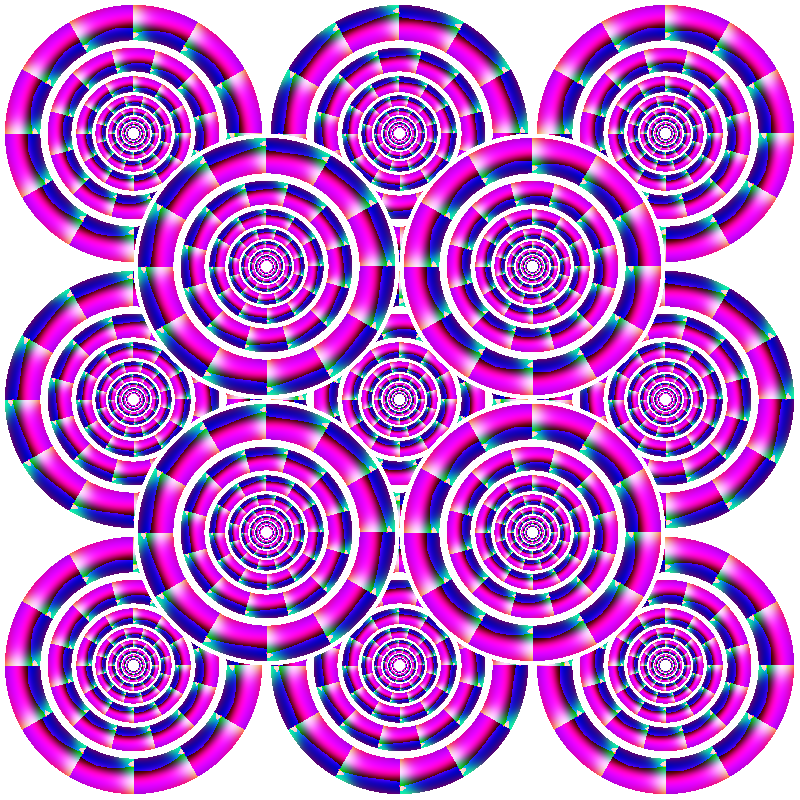}
    \includegraphics{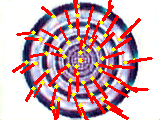}
     \caption{\textbf{EIGen-generated image} referred to as ``AI color Radial 2" in the perceptual experiment, and its calculated visual flow.}
    \label{fig:app:6}
\end{figure}

\begin{figure}
    \centering
    \includegraphics[width=1.0\linewidth]{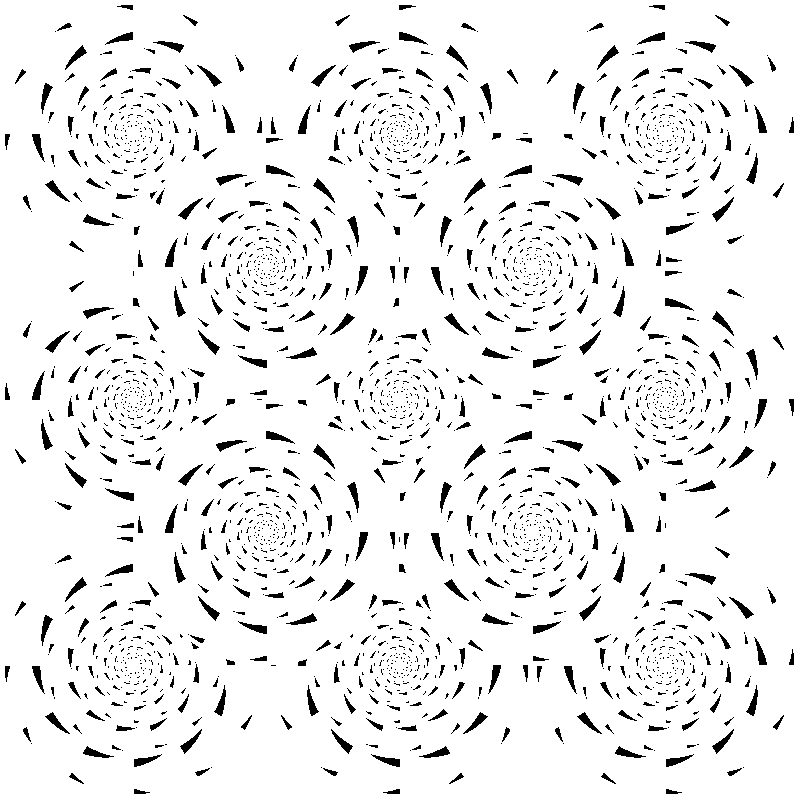}
    \includegraphics{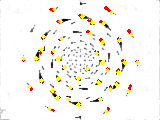}
     \caption{\textbf{EIGen-generated image} referred to as ``AI Medaka" in the perceptual experiment, and its calculated visual flow.}
    \label{fig:app:7}
\end{figure}

\begin{figure}
    \centering
    \includegraphics[width=1.0\linewidth]{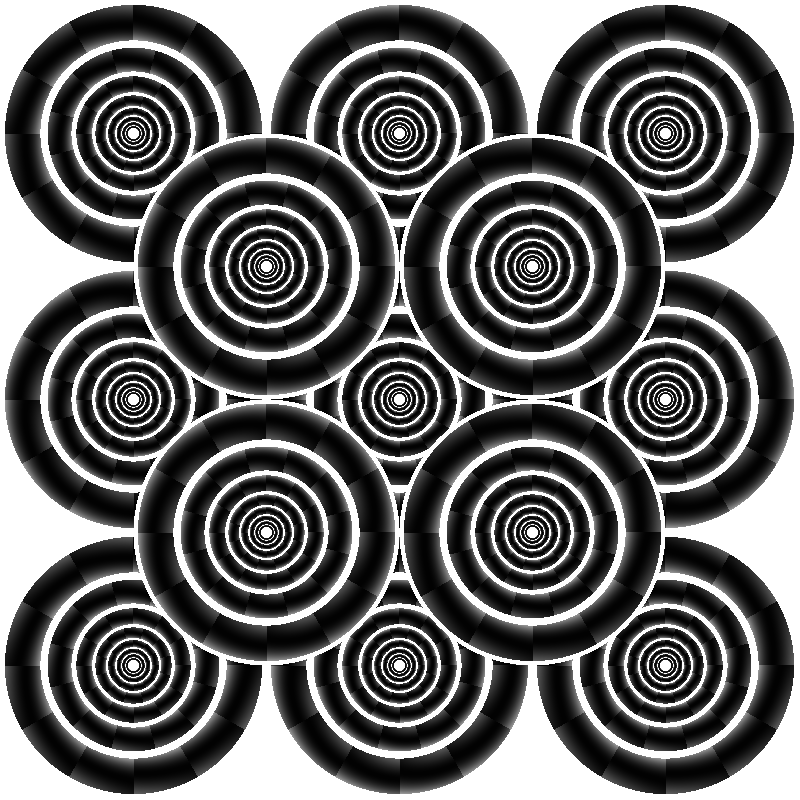}
    \includegraphics{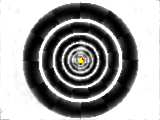}
     \caption{\textbf{EIGen-generated image} referred to as ``AI Control" in the perceptual experiment, and its calculated visual flow.}
    \label{fig:app:8}
\end{figure}

\end{document}